\documentclass[letterpaper, 10 pt, conference]{ieeeconf}  

\IEEEoverridecommandlockouts                              

\usepackage{amsthm}
\usepackage{graphicx}
\usepackage{caption}
\usepackage{subcaption}
\usepackage[linesnumbered,ruled,vlined]{algorithm2e}
\usepackage{algorithmicx}
\usepackage{amsmath}
\theoremstyle{plain}
\usepackage{mathtools}
\usepackage{epsfig}
\usepackage{amssymb}
\usepackage{epstopdf}
\usepackage{multicol}
\usepackage{multirow}
\usepackage{array}
\usepackage{url}
\usepackage{color}
\usepackage{float}
\usepackage{wrapfig}
\usepackage{placeins}
\usepackage{dblfloatfix}
\usepackage{tabu}
\usepackage{xcolor}
\usepackage{tabularx}
\usepackage{arydshln}
\usepackage[T1]{fontenc}
\title{\LARGE \bf
Neural Manipulation Planning on Constraint Manifolds
}

\author{Ahmed H. Qureshi, Jiangeng Dong, Austin Choe, and Michael C. Yip%
\thanks{A. H. Qureshi, J. Dong, A. Choe, and M. C. Yip are affiliated with the Department of Electrical and Computer Engineering, University of California San Diego, USA. J. Dong and A. Choe contributed equally as second authors. {\tt\small \{a1qureshi, jid103, achoe, yip\}@ucsd.edu}.} %
}

\begin{document}

\maketitle

\begin{abstract}
The presence of task constraints imposes a significant challenge to motion planning. Despite all recent advancements, existing algorithms are still computationally expensive for most planning problems. In this paper, we present Constrained Motion Planning Networks (CoMPNet), the first neural planner for multimodal kinematic constraints. Our approach comprises the following components: i) constraint and environment perception encoders; ii) neural robot configuration generator that outputs configurations on/near the constraint manifold(s), and iii) a bidirectional planning algorithm that takes the generated configurations to create a feasible robot motion trajectory. We show that CoMPNet solves practical motion planning tasks involving both unconstrained and constrained problems. Furthermore, it generalizes to new unseen locations of the objects, i.e., not seen during training, in the given environments with high success rates. When compared to the state-of-the-art constrained motion planning algorithms, CoMPNet outperforms by order of magnitude improvement in computational speed with a significantly lower variance.
\end{abstract}

\section{INTRODUCTION}
Efficient and scalable manipulation planning is of paramount importance in robotics and automation to solve real-world tasks. However, in most cases, manipulation of objects imposes hard kinematic constraints on the robot that limit its allowable range of motion. Examples of such instances include moving an object from one place to another that might also contain orientation constraints \cite{ott2006humanoid}, maintaining contact with the environment such as in robot surgery \cite{ballantyne2003vinci}, and performing bi-manual manipulation \cite{tsai1999robot}. In all of these cases, kinematic constraints form one or more low-dimensional manifolds of a set of robot configurations embedded in a high-dimensional ambient space, and are often known as manifold constraints \cite{kingston2018sampling}.

\begin{figure}
\vspace*{0.05in}
    \centering
       \includegraphics[width=8.5cm]{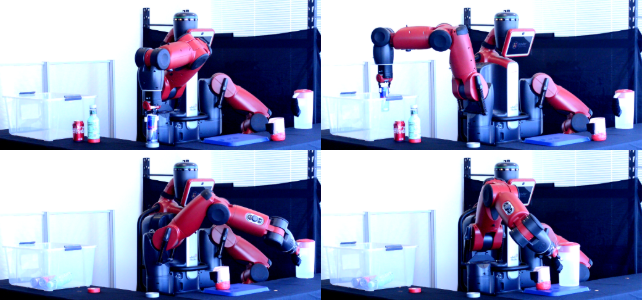}
    \caption{CoMPNet applied to a real 7DOF Baxter robot manipulator in the bartender environment. The task is to place the bottles and cans to the trash, and carefully move (without tilting) the mug and pitcher to the tray. The top and bottom rows show the start and goal states at the beginning and end of this table cleaning task.
}\label{bax_real}
\vspace*{-0.1in}\end{figure} 

Sampling-based Motion Planning (SMP) algorithms have emerged as a standard tool in robotics that find collision-free paths between the given states by randomly sampling the robot configuration space \cite{lavalle2006planning}. However, incorporating various kinematic constraints into the SMP algorithms is challenging as the underlying constraint manifold is usually of zero-measure. Therefore, the probability of generating samples on the constraint manifold by randomly sampling robot joint-values/configurations is not just low but zero \cite{kingston2018sampling}. Recently, SMP methods have been extended to plan under manifold constraints on various challenging problems \cite{kingston2018sampling}. However, despite these advancements, the existing techniques are computationally inefficient and therefore, frequently impractical for real-world manipulation tasks.

Recent developments also lead to imitation-based planners that learn to plan by imitating an oracle planner \cite{ichter2018learning}\cite{qureshi2018deeply}\cite{bency2019neural}\cite{ichter2019robot}\cite{qureshi2019motion}. These planners are known for their extremely fast computational speed during online planning. Motion Planning Networks (MPNet) \cite{qureshi2019motion}\cite{qureshi2019motionb} is one of the learning-based planners that can generate end-to-end collision-free paths and is also combined with SMPs for worst-case theoretical guarantees. MPNet is a deep neural networks-based bidirectional iterative planning algorithm and it is shown to demonstrate consistently better performance than state-of-the-art SMP methods, e.g., \cite{gammell2015batch}, in challenging motion planning problems. However, MPNet, along with extensions to it, only consider unconstrained planning problems.

In this paper, we propose Constrained Motion Planning Networks (CoMPNet)\footnote{Supplementary material and video demonstrations are available at https://sites.google.com/view/constrainedmpnet/home} that extends MPNet to plan under multimodal kinematic and task-specific constraints. Our proposed framework is a full-stack, computationally-efficient planner that solves motion planning problems for both reach and manipulation tasks, i.e., reaching to grasp arbitrary object and manipulating the grasped objects under various kinematic constraints. To the best of our knowledge, CoMPNet is the first learning-based planning algorithm that finds feasible paths under multiple hard kinematic constraints. CoMPNet comprises the observation (environment perception) and task (constraint) encoders whose outputs are given to the neural planning network that, together with the bidirectional planning algorithm, generates a feasible path on the constraint manifolds between the given start and goal configurations. We evaluate our method on challenging tasks that include both simulations and a real-robot setup (Fig. \ref{bax_real}). Our results show that CoMPNet outperforms existing methods in terms of computation speed and generalizes to new planning problems outside its training demonstrations with high success rates.

\section{Related Work}
In this section, we present a brief overview on existing non-sampling- and sampling-based approaches specifically addressing constraint motion planning (CMP), which represent a very challenging subset of planning problems with wide applications to robotics.

In non-sampling-based planning methods, one of the prominent tools is Trajectory Optimization (TrajOpt) \cite{ratliff2009chomp}\cite{schulman2014motion}. TrajOpt relaxes the hard constraints into soft constraints by defining them as penalty functions and optimizing them over the entire trajectory to find a motion plan. Due to this relaxation, TrajOpt methods weakly satisfy the given constraints and do poorly on long-horizon problems. Bonalli et al. \cite{bonalli2019trajectory} extends TrajOpt to implicitly-defined constraint manifolds. However, their performance in challenging robotics problems is yet to be explored and analyzed.

In sampling-based planning approaches, the multi-query Probabilistic Road Maps (PRMs) \cite{kavraki1998probabilistic} and single-query Rapidly-exploring Random Trees (RRTs) \cite{lavalle1998rapidly} and their variants \cite{karaman2011sampling} are widely known. These methods were initially devised for unconstrained problems such as collision avoidance, and incorporating kinematic constraints into them is challenging due to zero probability of sampling constraint satisfying configurations \cite{kingston2018sampling}. To address this challenge, there exist projection- and continuation-based strategies that generate samples on the constraint manifolds for SMP methods.

The projection-based methods project the given configurations to the constraint manifolds using the constraint equations. Typically, projections are performed using Inverse Kinematics (IK)-based iterative solvers that employ Jacobian (pseudo-) inverses of the robot model. These projection-based approaches have been used for special cases such as closed-chain kinematic constraints \cite{xie2004kinematics} as well as for general end-effector constraints \cite{stilman2010global}. In a similar vein, Berenson. et al. \cite{berenson2011task} proposed CBiRRT, a bidirectional RRT planner that uses general constraint representation known as Task Space Regions (TSRs) together with a Jacobian pseudo-inverse projection operator for CMP.

The continuation-based methods compute a tangent space from a known constraint-satisfying configuration to locally parametrize the underlying constraint manifold. The new constraint-satisfying configurations and local motions are generated by projecting the configurations sampled from the computed tangent space. In \cite{stilman2010global}, the idea of continuation has been used to sample the neighborhood of a constraint-satisfying configuration, and then project them to the constraint manifold. Recent advancements also led to bidirectional RRT-based algorithms called Atlas-RRT \cite{jaillet2017path} and Tangent Bundle (TB)-RRT \cite{kim2016tangent} that compose tangent spaces into an atlas instead of discarding them to represent the constraint manifold. Atlas-RRT computes half-spaces to separate tangent spaces into tangent polytypes for uniform coverage. In contrast, TB-RRT lazily evaluates configurations for constraint adherence and does not separate tangent spaces, leading to overlap and sometimes invalid states.

A very recent work called Implicit MAnifold Configuration Space (IMACS) \cite{kingston2019exploring} decouples constraint adherence methods like projection and continuation from the choice of underlying motion planning algorithms. Although their approach allows a broad range of SMP algorithms to operate under kinematic constraints, ultimately, a classic bidirectional RRT is preferred due to the poor computational speed of advance methods like RRT* \cite{karaman2011sampling} and its variants \cite{gammell2015batch,qureshi2015intelligent,gammell2014informed} in CMP.

In contrast to the methods mentioned above, CoMPNet leverages past planning experiences for learning a deep neural model and generates samples on the implicit manifolds during execution to quickly find a path solution for both constrained and unconstrained planning problems. Furthermore, our approach can also be combined with existing sampling-based CMP methods for worst-case theoretical guarantees while still retaining the computational benefits.
\begin{figure*}
\vspace*{-0.2in}
    \centering
       \includegraphics[width=16.5cm, height=10.0cm]{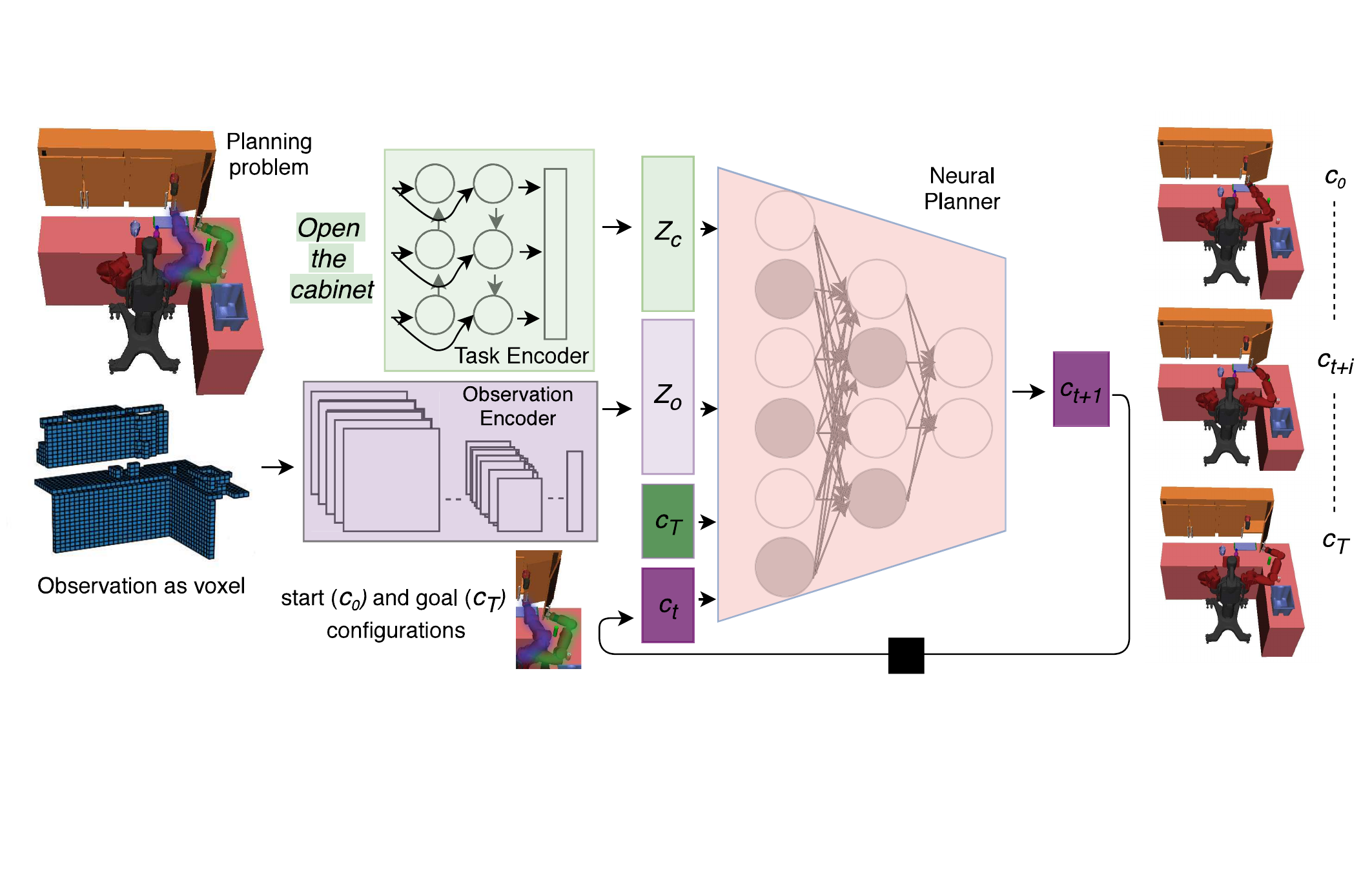}\vspace*{-0.8in}
    \caption{CoMPNet generating configurations for the door opening task. Our neural planner takes task encoding $Z_c$, observation encoding $Z_o$, start configuration $c_0$ (purple) and goal $c_T$ (green) configuration as input, and incrementally generates intermediate configurations $c_{t+i}$ for the path planning.  
}\label{bax_exe}
\vspace*{-0.1in}\end{figure*}   
\section{Problem Definition}
Let $\mathcal{C} \in \mathbb{R}^d$ be a $d$-dimensional configuration space (C-space) where $c \in \mathcal{C}$ represents a robot's configuration such as joint angles. Since the robot surroundings usually contain obstacles, the C-space comprises obstacle $(\mathcal{C}_{obs})$ and obstacle-free $(\mathcal{C}_{free}=\mathcal{C}\backslash\mathcal{C}_{obs})$ spaces. In general, the aim of motion planning is to find a continuous path $\sigma \subset \mathcal{C}_{free}$ that lies in obstacle-free space and connects the given start $(c_{init})$ and goal $(c_{goal})$ configurations of the robot. In CMP, the challenge is not just to avoid collisions but also to find a path that satisfies the given constraint function $F(c: \mathcal{C} \mapsto \mathbb{R}^k)$, where $k$ is the number constraints to be imposed. A configuration $c$ is said to satisfy the given constraint function if $F(c)=0$. Therefore, the constraint function defines a $(d-k)$-dimensional manifold within C-space, i.e., $\mathcal{M}=\{c\in\mathcal{C}| F(c)=0\}$. Hence, the goal of CMP is to find a continuous feasible path $\sigma \subset \mathcal{M}_{free}$ connecting the given start and goal configurations, where $\mathcal{M}_{free}= \mathcal{M} \cap \mathcal{C}_{free}$.
\begin{algorithm}[h]
\DontPrintSemicolon 
$T_a \gets c_{init}$; $c_t \gets c_{init}$\;
$T_b \gets c_{goal}$; $c_T \gets c_{goal}$\;
\For{$i \gets 0$ \textbf{to} $n$} {
$c^a_{t+1} \gets \mathrm{NeuralConfig}(Z_c,Z_o,c_t,c_T)$\;
$c^a_{near} \gets \mathrm{NearestNode}(c^a_{t+1},T_a)$\;
$c^{a}_{new} \gets \mathrm{TransverseManifold}(c^a_{t+1},c^a_{near},T_a)$\;
$c^b_{near} \gets \mathrm{NearestNode}(c^{a}_{new},T_b)$\;
$c^{b}_{new} \gets \mathrm{TransverseManifold}(c^{a}_{new},c^b_{near},T_b)$\;

\If{$\mathrm{Reached}(c^{a}_{new},c^{b}_{new})$}
   {
   \Return{$\mathrm{ExtractPath}(T_a,c^{a}_{new},c^{b}_{new},T_b)$}\;
   	  
   }
 \Else
 {
    $c_t \gets c^{a}_{new} $; $c_T \gets c^{b}_{new} $\;
	$\mathrm{Swap}(T_a,T_b)$\;
	$\mathrm{Swap}(c_t,c_T)$\;
 }

   }
\Return{$\varnothing$}\;
\caption{CoMPNet($Z_c,Z_o,c_{init},c_{goal}$)}
\label{algo:1}
\end{algorithm}
\section{Neural Manipulation Planning}
In this section, we formally present our novel constraint motion planning framework called CoMPNet that comprises the following components:

\subsection{Task Encoder}
The task encoder takes the task specification as the input for encoding. The task specification in our case is a text description, such as \textit{``carefully move mug to the tray''}, \textit{``open the cabinet''}, etc., that encapsulates the underlying constraints. For instance, the word \textit{``carefully''} in the task specification \textit{``carefully move mug to the tray''} implies both stability and collision-avoidance constraints. The output of the task encoder is a fixed size latent space encoding $Z_c \in \mathbb{R}^{d_1}$ with dimensionality $d_1$. In our task encoder, we obtain the task specification representations using a pretrained recurrent-neural network-based model called InferSent \cite{conneau2017supervised}. We further process these representations by a feed-forward neural network, which is trained with other CoMPNet modules, to get an embedding of size $d_1$. 

\begin{algorithm}[h]
\DontPrintSemicolon 
$c_0 \gets c_{near}$\;
$c_1 \gets c_{near}$\;
$i=0$\;
\While{$\|c_i-c_{target}\|_2>\varepsilon$} {
$c_{i+1} \gets \mathrm{Proj} (c_i+\Delta s(c_{target}-c_i))$\;
\If{$\mathrm{CollisionFree} (c_i,c_{i+1})$}
   {
   $T.\mathrm{InsertNode} (c_{i+1})$\;
   $T.\mathrm{InsertEdge} (c_i,c_{i+1})$\;
   $i=i+1$\;	  
   }
 \Else
 {
	 \Return{$c_{i}$}\;
 }

   }
\caption{TransverseManifold($c_{target},c_{near},T$)}
\label{algo:2}
\end{algorithm}

\subsection{Observation Encoder}
The observation encoder takes the environment perception information as an input to embed them into a latent space $Z_o \in \mathbb{R}^{d_2}$ of dimension $d_2$. In our settings, we obtain environment perception as a 3D point-cloud depth data, which is quantized as a voxel grid via voxelization. Ideally, 3D convolutional neural networks (CNNs) are used to process voxel-grids. However, in practice, 3D-CNNs are computationally expensive and impractical for large point-cloud data as their representations are inherently cubic and contain empty volume.  Therefore, we convert the voxel grid of dimension $L\times W \times H \times C$, where $C$ is a number of channels, to voxel image with voxel patches of dimension $L\times W \times (HC)$ and use 2D-CNNs to process them (for more details, refer to \cite{zhang2018efficient}).

\subsection{Planning Network}
It is a stochastic feed-forward neural network that drives its stochasticity from Dropout \cite{srivastava2014dropout} during execution. The planning network (PNet) takes the observation encoding $Z_o$, task encoding $Z_c$, robot current $c_t$, and goal $c_T$ configurations as input and learns to generate the next configuration $c_{t+1}$ that will bring the robot closer to the goal configuration on the constraint manifold. Since PNet predicts one step at a time, the path is formed incrementally (Fig. \ref{bax_exe}).
\subsection{Training Objective}
We train the task encoder's feed-forward neural network, observation encoder network, and planning network end-to-end through supervised imitation learning. An oracle planner is used to generate demonstration trajectories for a given set of constrained planning problems. A demonstration trajectory comprises waypoints $\sigma=\{c^*_0,\cdots,c^*_T\}$, where $c^*_0$ and $c^*_T$ correspond to given start and goal configurations, respectively. Our training objective is to optimize the mean-square error between the predicted configurations and the actual configurations from the training data, i.e.,
\begin{equation}\vspace*{-0.1in}\label{mse}
\cfrac{1}{N_B} \sum^{N}_{i=0} \sum^{T_i-1}_{j=0} ||c_{i,j+1}-c^*_{i,j+1}||^2,
\end{equation}
where $T_i$ is the length of each given path, $N \in \mathbb{N}$ is the number of paths in the training batch, and $N_B$ is an averaging term. To train all modules together, we backpropagate the gradient of the loss function (Eqn. \ref{mse}) from the planning network to the task and observation encoders.
\begin{algorithm}[b]
\DontPrintSemicolon 
\For{$i \gets 0$ \textbf{to} $n$} {
$\Delta x \gets F(c)$ 

\If{$\Delta x <\varepsilon$}
   {
   \Return{$c$}\;
   	  
   }
 \Else
 {
    $J \gets \mathrm{GetJacobian}(c)$\;
	$J^+ \gets \mathrm{Get PseudoInverse}(J)$\;
	$\Delta c \gets J^+ \Delta x$\;
	$c \gets (c-\Delta c)$\;
 }

   }
\caption{Proj($c$)}
\label{algo:3}
\end{algorithm}

\subsection{Bidirectional Neural Planning Algorithm}
Algorithm \ref{algo:1} outlines our bidirectional neural planning algorithm, whereas its essential components and overall execution are described as follows.
\subsubsection{NeuralConfig}
The NeuralConfig function uses the planning network to generate neural configuration on/near the manifold by taking the current $c_t$ and desired goal $c_T$ configurations together with the task $Z_c$ and observation $Z_o$ encodings as the input (Fig. \ref{bax_exe}), i.e.,
\begin{equation*}
c_{t+1} \gets \mathrm{PNet}(Z_c,Z_o,c_t,c_T)
\end{equation*}
In NeuralConfig, the given configurations are normalized to $[-1,1]$ before passing to the neural network and the generated neural configurations are unnormalized to robot joint-space for other path planning routines.
\subsubsection{Transverse Manifold}
The TransverseManifold function (Algorithm \ref{algo:2}), also known as geodesic interpolation, extends the given $T$ from the node $c_{near}$ towards the target node $c_{target}$ on the constraint manifold in small adaptive steps $\Delta s \in \mathbb{R}$. The procedure takes a linear $\Delta s$ step towards the given target and project it to the constraint manifold using the projection operator (Algorithm \ref{algo:3}). A projection operator (Proj) takes the configuration $c$ and project it to the constraint manifold via gradient descent using inverses or pseudo-inverses of the Jacobian $J(c)$ of the differentiable constraint function $F$. We define $F$ using TSRs \cite{berenson2011task} that returns displacement $\Delta x$ in task space. The iterative projection finds, if one exists in a given loop limit, a constraint-satisfying configuration $c$ such that $F(c)< \varepsilon$. The geodesic extension using Proj continues until it reaches $c_{target}$ or if the collision happens.
\subsubsection{Algorithm summary}
Let $T_a$ and $T_b$  contain the waypoints and their connecting edges from start to the goal configuration and from goal to the start configuration, respectively. The planning (Algorithm \ref{algo:1}) begins by generating a neural configuration $c^a_{t+1}$ towards the goal $c_T$ followed by finding its nearest node $c^a_{near}$ within $T_a$. The transverse manifold extends $T_a$ from $c^a_{near}$ towards $c^a_{t+1}$ on the manifold and stops before collision leading to $c^a_{new}$. The algorithm then proceeds by finding the nearest node $c^b_{near}$ to $c^a_{new}$ within $T_b$. The transverse manifold function extends $T_b$ from $c^b_{near}$ towards $c^a_{new}$ and terminates before collision occurs by producing $c^b_{new}$. If $c^a_{new}$ and $c^b_{new}$ reach each other, the algorithm extracts the end-to-end collision-free path connecting given start and goal configurations on the constraint manifold. The extracted path is further processed via smoothing to remove any redundant states before returning it as a feasible path solution. In case the $c^a_{new}$ and $c^b_{new}$ do not meet, the procedure continues by updating $c_t$ with $c^a_{new}$ and $c_T$ with $c^b_{new}$ followed by swapping the roles of $(T_a,c_t)$ with $(T_b,c_T)$ to solicit bidirectional path generation. 

Note that our framework intelligently uses the trained planning network to generate configurations bidirectionally, i.e.,
from given start to goal configuration and from given goal
to start configuration. It also computes nearest neighbors of the newly generated samples from each path to further solicit bidirectional geodesic extension on the constraint manifolds during each planning iteration. This is in contrast to MPNet algorithm that extends a path from $c_t$ to $c_{t+1}$ rather than from its nearest neighbors and rely on re-planning, making it less suitable for geodesic interpolation due to non-euclidean geometry of implicit manifolds.

\section{Implementation details}
We train CoMPNet neural models using PyTorch and port them to C++ via TorchScript for planning algorithm. The environments were set up in OpenRave, and for benchmark algorithms, we use their standard C++ implementations. The rest of the section provides details on the data generation, scene setup, and constraint representations. For more details, please refer to our supplementary material. 
\subsection{Data generation}
In this section, we describe the data generation pipeline from setting scenarios to obtaining training data. Tasks are designed to involve grasping objects and controlling orientations on a series of sequences. Both grasped objects and forced orientation impose constraint manifolds in the robot configuration space that the planners must satisfy in addition to reaching its sequence of goals. 
\begin{figure*}[h]
    \centering
    \begin{subfigure}[b]{0.327\textwidth}
     \includegraphics[width=6.0cm]{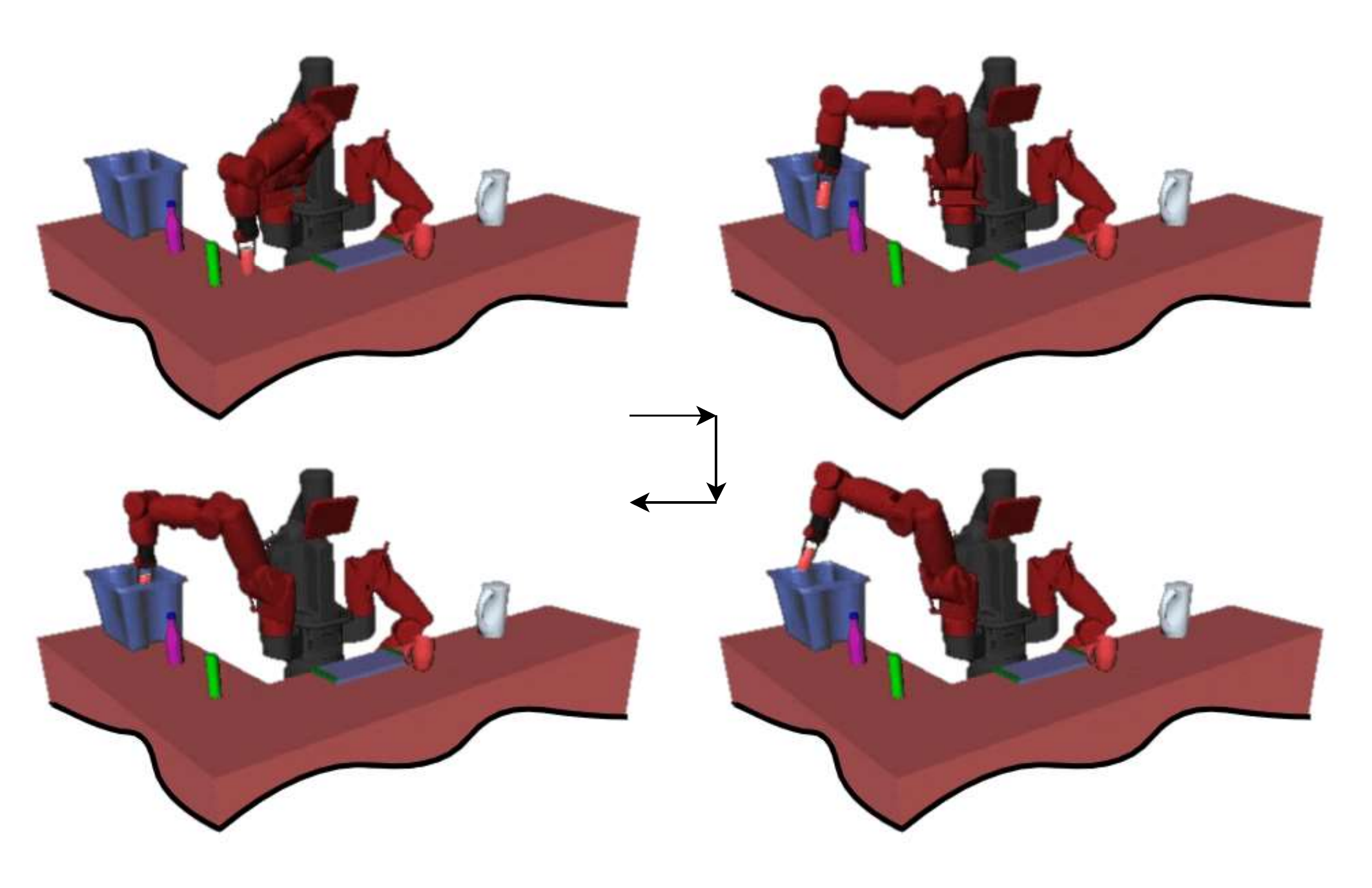}
       \caption{Move soda can to the trash.}
    \end{subfigure}
    \begin{subfigure}[b]{0.327\textwidth}
       \includegraphics[width=6.0cm]{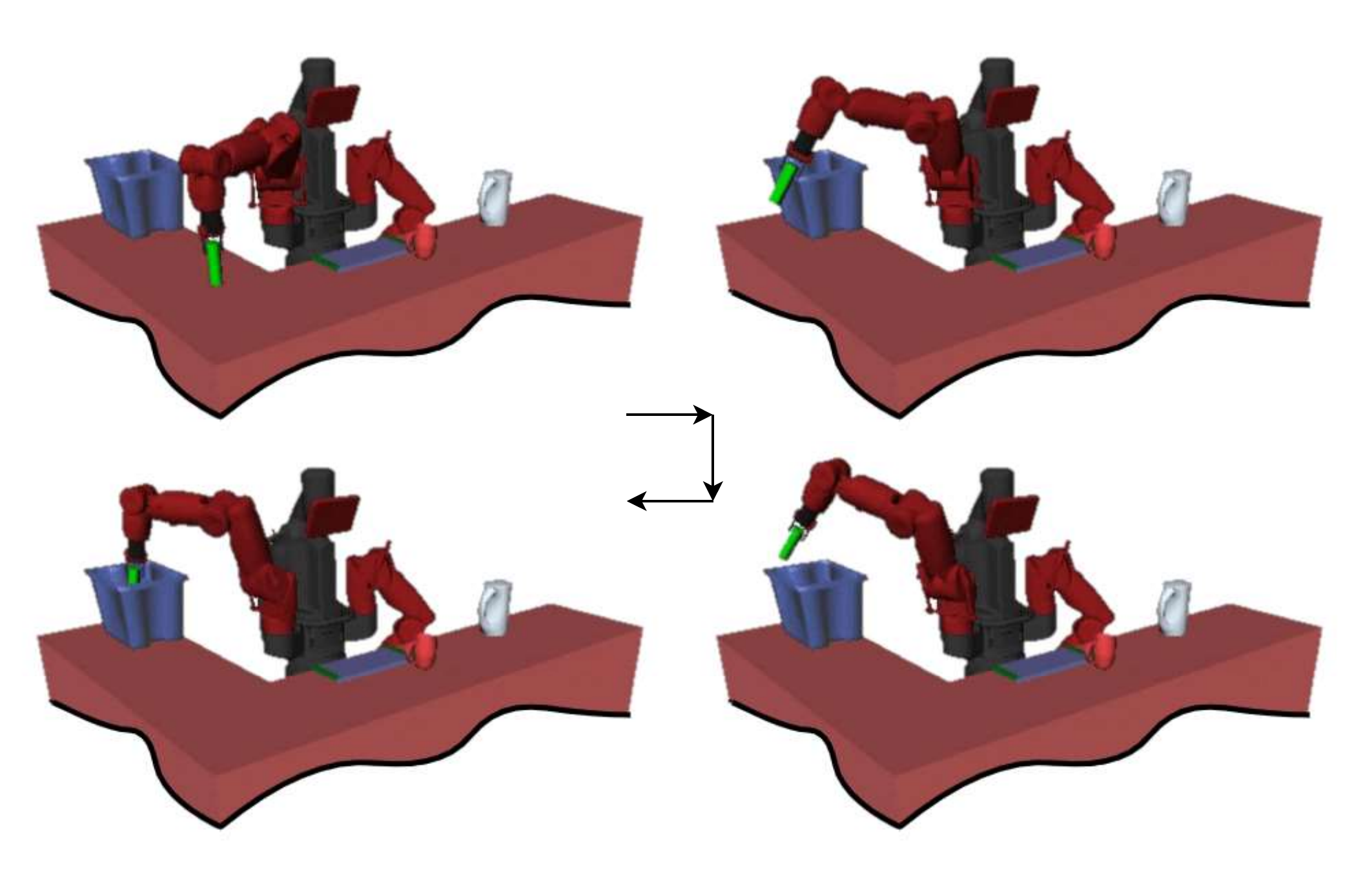}
        \caption{Move green can to the trash.}
    \end{subfigure}
    \begin{subfigure}[b]{0.327\textwidth}
       \includegraphics[width=6.0cm]{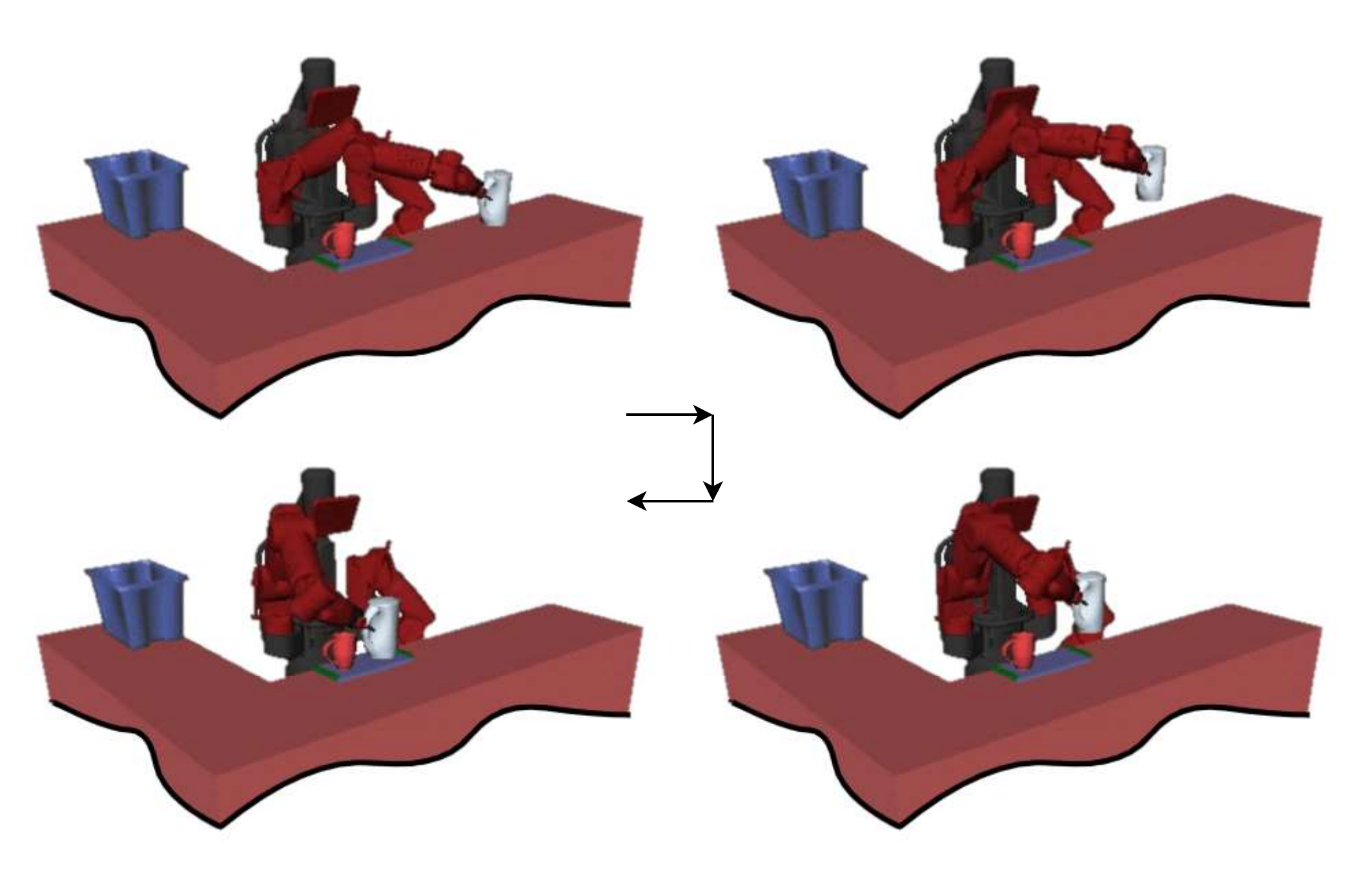}
        \caption{Carefully move kettle to the tray.}
    \end{subfigure}    
    \caption{Bartender setup: It requires the robot to clean the tables by placing the cans and bottle to the trash, and carefully moving the mug and kettle to the tray. Figs. (a-c) show some of the example subtasks.}\label{bartender}
\end{figure*}
\begin{figure}
\vspace*{-0.1in}
    \centering
       \includegraphics[width=9.0cm,trim={0cm 0cm 0cm 1.5cm},clip]{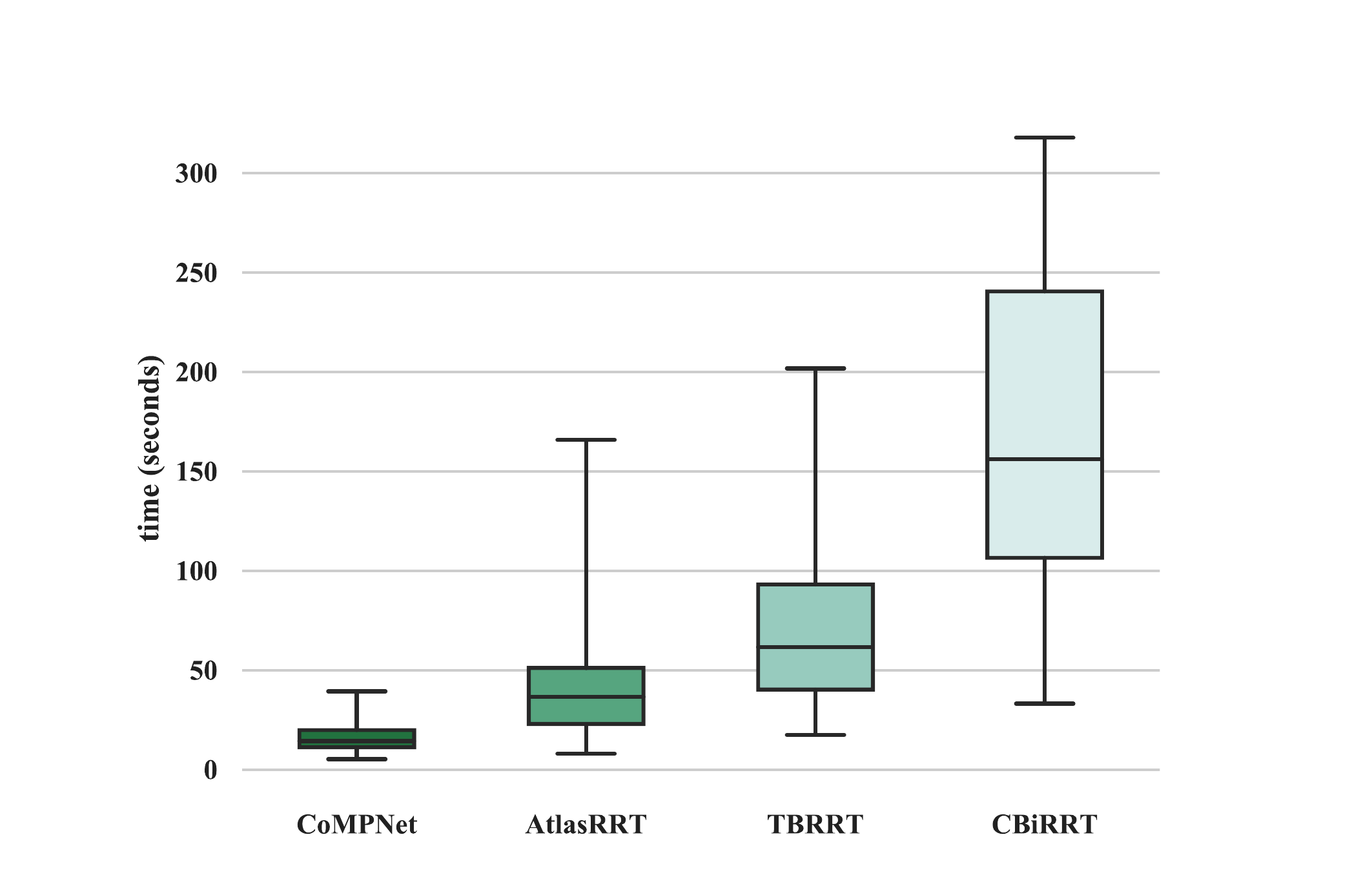}
    \caption{The plot shows the interquartile range, minimum and maximum of total computation time of presented methods to solve all manipulation problems in the bartender task. Note that, CoMPNet's computation time is consistent and significantly faster than other benchmark methods.
}\label{bartender_plot}
\vspace*{-0.25in}\end{figure}
\subsubsection{Scene setup}
We set up two practical scenarios, (i) bartender task (Fig. \ref{bartender}) and (ii) kitchen task (Fig. \ref{kitchen}), each of which includes multiple sub-tasks.

The bartender task involves five manipulatable objects comprising the soda can (red), juice can (green), fuze bottle (purple), red mug, and kettle. The task is to place the soda can, juice can, and fuze bottle to the trash bin, and carefully transfer (without tilting) the red mug and kettle to the tray. In this task, we created 30 unique environments by random placement of trash bin and tray, reachable by the the robot's right arm. For each environment, we further create 60-110 unique scenarios by placing the five manipulatable objects randomly on the table at the reacjable locations.  

The kitchen task involves seven manipulatable objects that included a soda can, juice can, fuze bottle, cabinet door, black mug, red mug, and pitcher. The task is to place the cans and bottle to the trash bin, open the cabinet to a given angle, carefully move (without tilting) the black and red mugs to the tray, and then move the pitcher into the cabinet. We create 60 unique environments by randomly placing a trash bin at the robot's right hand's reachable locations on the table and also by randomly setting the cabinet's door starting angle between $0$ to $\pi/3$. For each environment, we further create about 30 unique scenarios through the random placement of the soda can, juice can, fuze bottle, tray, and pitcher at the robot's (right arm) reachable poses on the tables. Note that the starting pose of the black and red mugs are fixed in all cases, i.e., inside the cabinet, whereas their goal poses are on the randomly placed tray. 
\subsubsection{Observation data}
From all generated scenarios, we get the point-cloud depth data using multiple Kinect sensors before solving each of the given sub-tasks in the order provided by the task planner. The point-cloud data from various sensors are stacked and converted to voxels via voxelization. The voxel dimensions for bartender and kitchen tasks/sub-tasks were $33\times 33\times 33$ and $32\times 32\times 32$, respectively. 

\subsubsection{Planning trajectories (training \& testing)}
We pull out more than $10\%$ of the randomly generated scenarios for testing. For the remaining scenes, we get demonstration trajectories using CBiRRT \cite{berenson2011task} and use them for training the CoMPNet's neural models. In the real robot setup (Fig. \ref{bax_real}), we replicate one of the bartender scenarios and execute CoMPNet's planned motion to demonstrate the transferability of our simulation experiments to the real-robot settings. 

\begin{figure*}[t]
    \centering
    \begin{subfigure}[b]{0.327\textwidth}
     \includegraphics[height=5.0cm]{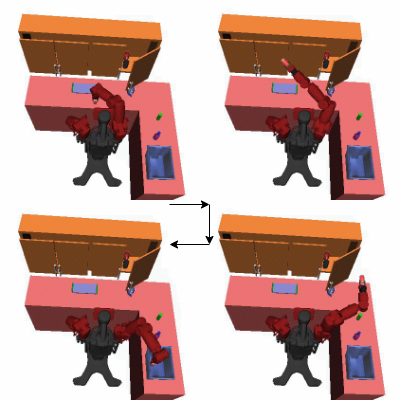}
       \caption{Move soda can to the trash.}
    \end{subfigure}
    \begin{subfigure}[b]{0.327\textwidth}
       \includegraphics[height=5.0cm]{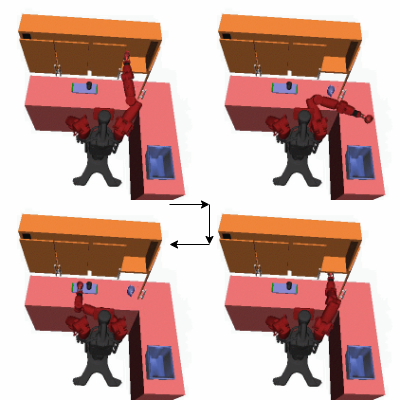}
        \caption{Carefully move red mug to the tray.}
    \end{subfigure}
    \begin{subfigure}[b]{0.327\textwidth}
       \includegraphics[height=5.0cm]{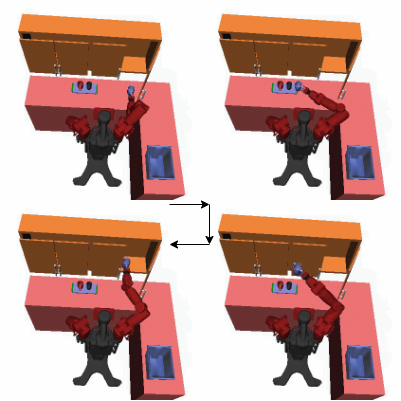}
        \caption{Carefully move pitcher to the cabinet.}
    \end{subfigure}    
    \caption{Kitchen setup: Figs. (a-c) and Fig. \ref{bax_exe} show some instances of the CoMPNet's path execution in these scenarios.}\label{kitchen}
\end{figure*}
\subsection{Constraint \& Task Representations}
We represent the constraint function $F$ using Task Space Regions (TSRs) \cite{berenson2011task} and our task encoder takes the constraint information as text encoding using InferSent \cite{conneau2017supervised}. We train our CoMPNet for both reach and manipulation tasks, and their text description as follows:\\ i) \textbf{Reach tasks:} The text description is \textit{``Reach the \underline{object name}''}. The \textit{\underline{object name}} includes \textit{``soda can''}, \textit{``juice can''}, \textit{``fuze bottle''}, \textit{``red mug''}, \textit{``black mug''}, \textit{``kettle''}, \textit{``pitcher''}, and \textit{``cabinet''} depending on the given setting (bartender or kitchen) and their task plan.\\ ii) \textbf{Manipulation tasks:} In the bartender scenarios, the text descriptions are \textit{``move the \underline{object name A} to the trash''} and \textit{``carefully move the \underline{object name B} to the tray''}. In the kitchen scenarios, the text description are \textit{``move the \underline{object name A} to the trash''}, \textit{``open the cabinet''}, \textit{``carefully move the \underline{object name C} to the tray''}, and \textit{``carefully move the pitcher to the cabinet''}. The \textit{\underline{object name A}} includes \textit{``soda can''}, \textit{``juice can''}, and \textit{``purple bottle''}. The \textit{\underline{object name B}} gets \textit{``red mug''}, and \textit{``kettle''}. The \textit{\underline{object name C}} corresponds to \textit{``red mug''}, and \textit{``black mug''}.
\begin{table*}[t]
\centering 
\begin{tabular}{cccccc}\hline
\multirow{2}{*}{Setup}&\multirow{2}{*}{Objects}&\multicolumn{4}{c}{Algorithms}\\\cline{3-6}
&&\multicolumn{1}{c}{CoMPNet}&\multicolumn{1}{c}{CBiRRT}&\multicolumn{1}{c}{Atlas-RRT}&\multicolumn{1}{c}{TB-RRT}\\\hline

\multirow{2}{*}{Bartender}&\multirow{1}{*}{J/F/S}& \multirow{1}{*}{$\boldsymbol{4.92 \pm 2.42}$} & \multirow{1}{*}{$54.81 \pm 25.82$}  &\multirow{1}{*}{$14.24 \pm 9.331$}&\multirow{1}{*}{$23.48 \pm 14.84$}\\ 
&\multirow{1}{*}{R/K}& \multirow{1}{*}{$\boldsymbol{1.17 \pm 0.93}$} & \multirow{1}{*}{$1.940 \pm 1.380$}  &\multirow{1}{*}{$1.276 \pm 0.361$}&\multirow{1}{*}{$1.574 \pm 5.641$}\\\cdashline{2-6}

\multirow{3}{*}{Kitchen}&\multirow{1}{*}{J/F/S}& \multirow{1}{*}{$\boldsymbol{4.28 \pm 2.86}$} & \multirow{1}{*}{$32.65 \pm 22.40$}  &\multirow{1}{*}{ $24.87 \pm 19.82$}&\multirow{1}{*}{$27.55 \pm 21.47$}\\
&\multirow{1}{*}{C}& \multirow{1}{*}{$\boldsymbol{0.03 \pm 0.02}$} & \multirow{1}{*}{$00.05 \pm 00.04$}  &\multirow{1}{*}{ $00.04 \pm 00.03$}&\multirow{1}{*}{$00.05 \pm 00.04$}\\
&\multirow{1}{*}{R/B/P}& \multirow{1}{*}{$\boldsymbol{9.16 \pm 3.04}$} & \multirow{1}{*}{$49.79 \pm 22.95$}  &\multirow{1}{*}{ $41.28 \pm 24.02$}&\multirow{1}{*}{$46.61 \pm 26.07$}
 \\ \hline
\end{tabular}
\caption{The mean computation times with standard deviations of solving manipulation problem of each object, grouped by their constraint types, in both bartender and kitchen environments. The objects are denoted by their first letter. It can be seen that CoMPNet computation times are lower and more consistent across different problems than other methods.} \label{tab}
\end{table*}
\begin{figure}
\vspace*{0.05in}
    \centering
       \includegraphics[width=9.0cm, trim={0cm 0cm 0cm 1.5cm},clip]{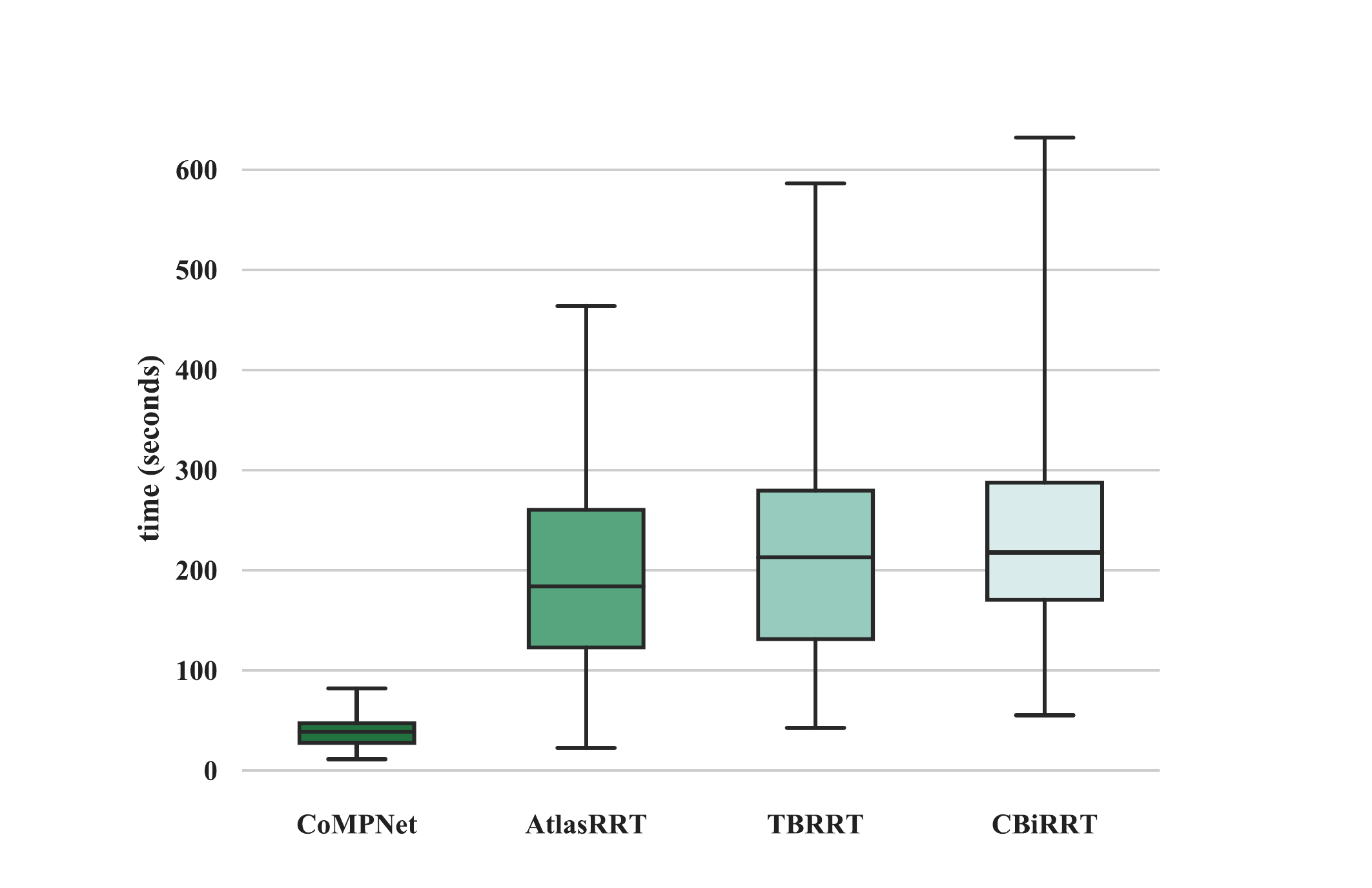}
    \caption{The plot exhibits the interquartile range, minimum and maximum of total computation time of CoMPNet, Atlas-RRT, TB-RRT, and CBiRRT to solve all manipulation tasks in the kitchen setup. Note that, CoMPNet's has significantly narrower and lower ranges of computational time than other benchmark algorithms.}\label{kitchen_plot}
\vspace*{-0.5in}\end{figure}
\section{Results}
In this section, we present the computation time comparison of CoMPNet and state-of-the-art CMP methods named CBiRRT \cite{berenson2011task}, Atlas-RRT \cite{jaillet2017path}, and TB-RRT \cite{kim2016tangent} on several bartender and kitchen tasks. All experiments were performed on a system with 32GB RAM, GeForce GTX 1080
GPU, and 3.40GHz$\times$8 Intel Core i7 processor. 
\begin{table*}[b]
\centering 
\begin{tabular}{ccccc}\hline
\multirow{2}{*}{Tasks}&\multicolumn{3}{c}{Algorithms}\\\cline{2-5}
&\multicolumn{1}{c}{MPNet (with Proj)}&\multicolumn{1}{c}{CoMPNet (w/o task encoder)}&\multicolumn{1}{c}{CoMPNet (with one-hot)}&\multicolumn{1}{c}{CoMPNet}\\\hline

\multirow{1}{*}{Bartender}& \multirow{1}{*}{$18.80 \pm 8.90$ $(75.1\%)$} & \multirow{1}{*}{$18.84 \pm 8.96$ $(82.1\%)$}  &\multirow{1}{*}{$17.00 \pm 8.31$ $(87.8\%)$}&\multirow{1}{*}{$16.52 \pm 7.67$ $(88.3\%)$}\\ 

\multirow{1}{*}{Kitchen}& \multirow{1}{*}{$44.01 \pm 9.53$ $(68.6\%)$} & \multirow{1}{*}{$41.21 \pm 12.13$ $(86.2\%)$}  &\multirow{1}{*}{ $41.93 \pm 13.97$ $(89.7\%)$}&\multirow{1}{*}{$38.6 \pm 13.23$ $(91.1\%)$}\\ 
\end{tabular}
\caption{Ablation study: The total mean computation times with standard deviations and mean success rates are presented for CoMPNet and it's ablated models in solving all manipulation tasks in the bartender and kitchen environments.} \label{tab}
\end{table*}

\subsection{Comparative Studies}
In the unconstrained planning problems, i.e., reaching-to-the-target-object, we validate that CoMPNet's mean computation time is about 1-2 seconds, which is similar to results with MPNet \cite{qureshi2019motionb}, showing that performance improvements over gold-standard unconstrained SMPs are retained.

In the constrained manipulation planning problems of the bartender (Fig. \ref{bax_real} \& Fig. \ref{bartender}) and kitchen (Fig. \ref{kitchen}) environments, the mean success rates of all the presented methods were around $90\%$. Fig. \ref{bartender_plot} and Fig. \ref{kitchen_plot} compare algorithms in both environments using the box plots of mean accumulated computation times in solving all the manipulation tasks. Furthermore, Table I also provides the mean computation time with standard deviations of manipulating each object, grouped by their constraint types, in each setting.

Note that all test environments were randomly generated and were not seen by the CoMPNet during training. These environments are challenging, representing practical scenarios, and often requiring a planner to find convoluted long-horizon paths through narrow passages. For instance, Fig. \ref{kitchen}  (b) shows a CoMPNet path solution for manipulating a red mug in the kitchen setup. It is a non-trivial, long-horizon plan that transverses narrow passages formed by the door and other objects on the table.

Despite challenging planning problems, it can be seen that CoMPNet compared to other methods exhibits i) higher/similar success rates, ii) lower inter-quartile computational time ranges, iii) lower minimum and maximum computation times, and iv) lower mean computation times with a narrower standard-deviations. Although CoMPNet uses constraint adherence methods like classical CMP algorithms for traversing manifolds, its lower computation times indicate that the generated samples are mostly on the constraint manifold and does not rely on the constraint adherence operator significantly. We also observed that continuation-based operators are highly sensitive to their parameters, and lazy evaluation heuristic of TB-RRT often leads to invalid states causing poor performance than other methods. 

\subsection{Ablative Studies}
We present an ablation study to highlight the significance of the following components added to MPNet that led to CoMPNet for scalable CMP: 1) A projection operator for constraint adherence and steering on the manifold. 2) A planning algorithm with bidirectional constrained extensions to newly generated sample $c_{t+1}$ from its nearest neighbors $\{c^a_{near},c^b_{near}\}$ rather than its previous configuration $c_t$. 3) A task specification for scalability and multimodality which could be one-hot or text-based encodings. Hence, our first model is MPNet without re-planning phase and with a projection operator for steering. Second is the proposed CoMPNet framework without task encodings. Third and fourth are the proposed CoMPNet models with one-hot and text-based task specifications, respectively.

Table II presents the total mean computation times with standard deviations and success rates of all models mentioned above in the bartender and kitchen environments. Although MPNet performs well in unconstrained planning problems, it can be seen that MPNet with only a projection operator and no re-plannings performs poorly in terms of success rates than other models in CMP. Similarly, the task encoding also leads to improved success rates validating that it is a crucial component of CoMPNet. Furthermore, task specification such as one-hot or text-based representation gives similar performances. However, using a sequential embedding such as text-based constraint specification is vital as they allow scalability to an arbitrary number of constraint types. In contrast, other methods such as one-hot representations would scale poorly and become very limited in practice with a growing set of multi-task and multimodal constraints.
\section{Discussion}
In this section, we present a brief discussion on CoMPNet’s ability to solve multimodal constraint motion planning problems and exhibit probabilistic completeness on manifold coverage if merged with uniform C-space sampling methods.

\subsection{Multimodal Constraints}
To the best of authors' knowledge, CoMPNet is the first planning algorithm capable of handling multimodal constraints and exploring various constraint manifolds simultaneously. In the presented kitchen and bartender setups, the problems require: i) reaching to the target end-effector pose to grab the given object; ii) manipulating the grabbed object under various constraints such as stability and collision-avoidance; iii) opening the cabinet door, which also imposes constraints inculcated by the allowable rotation of the door's hinge. In all of these problems, CoMPNet's high success rate and low computation time validate that its planning network, conditioned on the observation and text-based task encodings, can implicitly transition between different constraint manifolds and can produce the constraint-adhering configurations efficiently.
\subsection{Stochasticity \& Manifold Coverage}
In this section, we highlight that CoMPNet, coupled with an exploration-based C-space sampling strategy, covers the constraint manifold leading to probabilistic completeness guarantees. The probabilistic completeness guarantees are that the planner will output a path solution, if one exists, with a probability of one, if it is allowed to run for a large number of iterations approaching infinity.

In CoMPNet, we apply Dropout \cite{srivastava2014dropout} with probability of $0.5$ to almost every layer of the planning network during offline training and online planning. The Dropout randomly skips the output of some of the neurons from its preceding neural network's layer according to the given probability $p \in [0,1]$. In \cite{gal2016dropout}, Yarin and Zubin use Dropout for uncertainty modeling in the neural networks. In our method, we use Dropout-based stochasticity in the planning network to generate configuration samples on/near the constraint manifold. The generated configuration samples that are slightly off are projected to the constraint manifold using the gradient-descent-based projection operator. In \cite{berenson2011task}\cite{kingston2019exploring}, it is proved that the projection operator (Algorithm \ref{algo:3}) combined with uniform C-space sampling fully explores the underlying constraint manifold with the running time approaching infinity. CoMPNet can also be combined with uniform C-space sampling techniques leading to the exploration-exploitation approach. The exploitation phase will be to leverage CoMPNet's planning network for a fixed number of iterations to generates samples on/near the constraint manifold, potentially leading to a path solution. The exploration phase will be to do uniform sampling after exploitation. Hence, the exploration-exploitation based sampling approach paired with a projection operator will cover the constraint manifold with probabilistic completeness, and the proof can be derived in the same way as in \cite{berenson2011task}\cite{kingston2019exploring}.

\section{Conclusions \& Future Works}
We proposed Constrained Motion Planning Networks (CoMPNet), a neural network-based bidirectional planning algorithm that is shown to solve complex planning problems under multimodal kinematic constraints in seconds and with significantly less variability than state-of-art methods that take minutes and have high variability. CoMPNet also encapsulates MPNet \cite{qureshi2019motion} as it not only solves constrained manipulation problems but also unconstrained planning problems, i.e., reach to the given robot end-effector poses to grasp the objects. Furthermore, we also show that, similar to MPNet \cite{qureshi2019motion}, CoMPNet generalizes to unseen tasks that were not in the training examples with a high success rate and probabilistic completeness.

In our future studies, we plan to extend CoMPNet to integrated task and motion planning by leveraging its fast computational speed for almost real-time motion reasoning. In addition,  we also aim to incorporate dynamical constraints to allow a computationally efficient kinodynamic path planning for practical, real-world problems from the high-dimensional observation data.
\section*{Acknowledgments}
We thank Dmitry Berenson for the insightful discussions on the proposed work and Frank Park for sharing their TBRRT implementations.

\bibliographystyle{IEEEtran}
\bibliography{reference}
\nocite{*}
\end{document}